\documentclass[10pt,twocolumn,letterpaper]{article}

\usepackage[pagenumbers]{wacv}      

\usepackage{multirow}

\definecolor{wacvblue}{rgb}{0.21,0.49,0.74}
\usepackage[pagebackref,breaklinks,colorlinks,allcolors=wacvblue]{hyperref}

\def\wacvPaperID{2199} 
\def\confName{WACV}
\def\confYear{2026}

\title{ExDDV: A New Dataset for Explainable Deepfake Detection in Video}

\author{Vlad Hondru$^1$, Eduard Hogea$^{2}$, Darian Onchi\c{s}$^2$, {Radu Tudor} Ionescu$^{1,}$\thanks{Corresponding author: \texttt{raducu.ionescu@gmail.com}.}\\
$^1$University of Bucharest, Romania, $^2$West University of Timi\c{s}oara, Romania\vspace{-0.3cm}
}

\begin{document}
\maketitle

\begin{abstract}
The ever growing realism and quality of generated videos makes it increasingly harder for humans to spot deepfake content, who need to rely more and more on automatic deepfake detectors. However, deepfake detectors are also prone to errors, and their decisions are not explainable, leaving humans vulnerable to deepfake-based fraud and misinformation. To this end, we introduce ExDDV, the first dataset and benchmark for \textbf{Ex}plainable \textbf{D}eepfake \textbf{D}etection in \textbf{V}ideo. ExDDV comprises around 5.4K real and deepfake videos that are manually annotated with text descriptions (to explain the artifacts) and clicks (to point out the artifacts). We evaluate a number of vision-language models on ExDDV, performing experiments with various fine-tuning and in-context learning strategies. Our results show that text and click supervision are both required to develop robust explainable models for deepfake videos, which are able to localize and describe the observed artifacts. Our novel dataset and code to reproduce the results are available at  \small{\url{https://github.com/vladhondru25/ExDDV}}.

\end{abstract}

\definecolor{difficulty_green}{RGB}{0,205,0}
\definecolor{difficulty_orange}{RGB}{255, 140, 0}
\definecolor{difficulty_red}{RGB}{255,0,0}

\section{Introduction}
\label{sec:intro}

Online fraud and misinformation based on deepfake videos reached unprecedented expansion rates in recent years. A recent forensics report suggests that identity fraud rates nearly doubled, showing a significant rise in the prevalence of deepfake videos between 2022 and 2024, from $29\%$ to $49\%$\footnote{\url{https://regulaforensics.com/news/deepfake-fraud-doubles-down/}}. This rise of deepfake content is primarily caused by recent advances in generative AI, especially with the emergence of highly capable diffusion models \cite{chen-AAAI-2024, liu-ACMMM-2024, tian-ECCV-2024, stypulkowski-WACV-2024, wang-arXiv-2024, yuang-arXiv-2024, xu-arXiv-2024, xu-NeurIPS-2025}. The high quality and realism of generated videos put online users in difficulty of telling the difference between real and fake content. In this context, humans can turn to state-of-the-art automatic deepfake detectors for help \cite{bonettini-ICPR-2021, wang-AAAI-2023, yan-arXiv-2024, montserrat-CVPR-2020, hu-AAAI-2022, liu-WACV-2023, gu-ECCV-2022, haliassos-CVPR-2022}. To come in handy, such models need to be robust and trustworthy, while also providing explainable decisions that would enable users to gain important insights regarding the kinds of artifacts they should look for in a video. Yet, \citet{croitoru-arXiv-2024} showed that state-of-the-art deepfake detectors fail to generalize to content generated by new generative models (not seen during training). Moreover, to the best of our knowledge, the task of generating textual descriptions to explain the artifacts observed in deepfake videos has not been explored so far.

\begin{figure}[!t]
    \centering
    \includegraphics[width=1.0\linewidth]{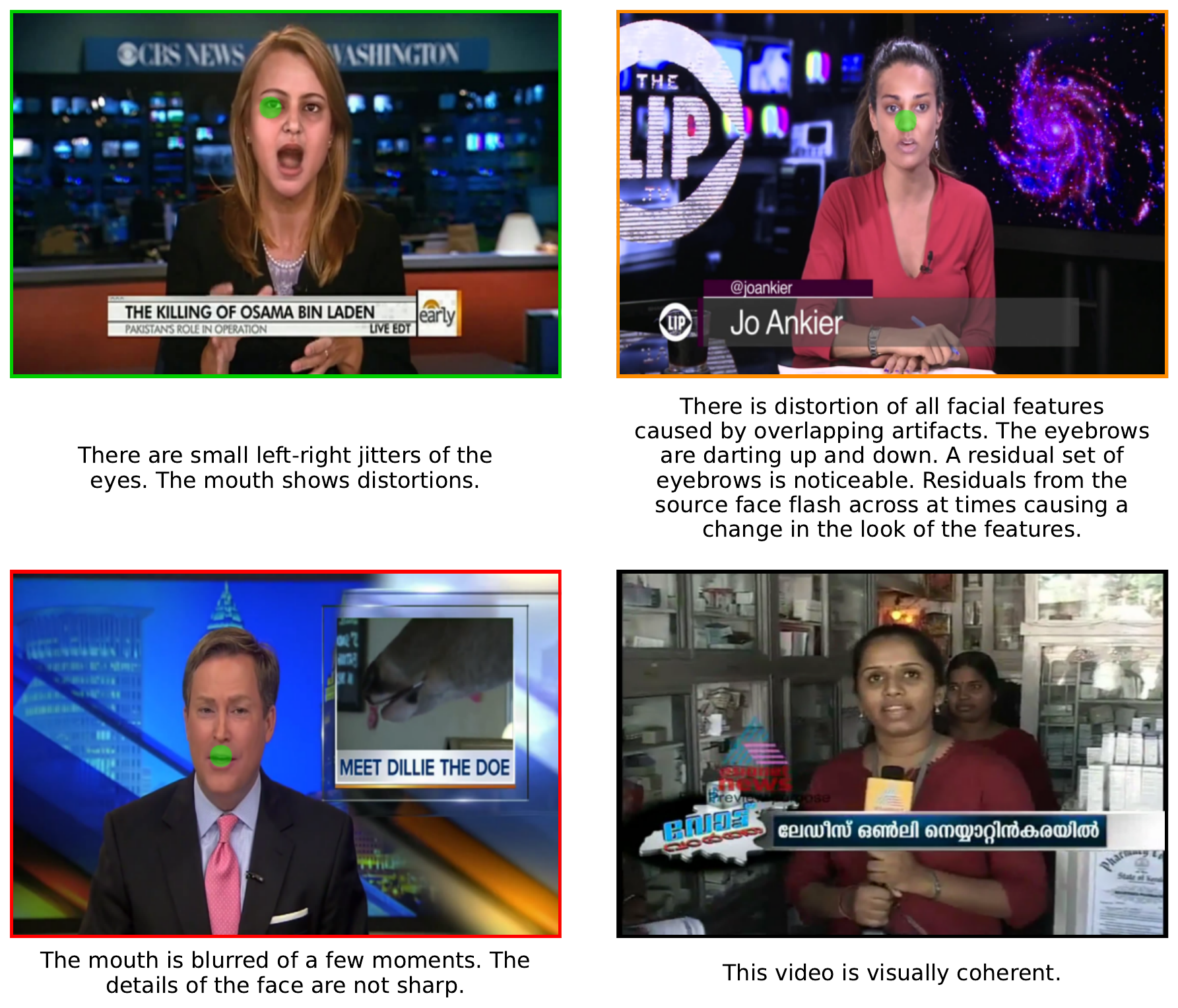}
    \vspace{-0.65cm}
    \caption{Examples of video frames from ExDDV with text and click annotations. Clicks are represented as large green dots. Real videos are not annotated with clicks or difficulty levels. The border color indicates the difficulty level: \textcolor{difficulty_green}{green}=easy, \textcolor{difficulty_orange}{orange}=medium, \textcolor{difficulty_red}{red}=hard, black=real.  Best viewed in color.}
    \label{fig:samples1}
    \vspace{-0.3cm}
\end{figure}

To this end, we introduce the first dataset and benchmark for \textbf{Ex}plainable \textbf{D}eepfake \textbf{D}etection in \textbf{V}ideo, called ExDDV. Our new dataset comprises approximately 5.4K real and deepfake videos that are manually annotated with text descriptions, clicks and difficulty levels. The text descriptions explain the artifacts observed by human annotators, while the clicks provide precise localizations of the described artifacts, as shown in \cref{fig:samples1}. The annotated videos are gathered from a broad set of existing datasets for video deepfake detection, including DeeperForensics \cite{jiang-CVPR-2020}, FaceForensics++ \cite{rossler-ICCV-2019}, DeepFake Detection Challenge \cite{dolhansky-arXiv-2020} and BioDeepAV \cite{croitoru-arXiv-2024}, to enhance the diversity of our collection. The inter-annotator agreement ($0.6238$ cosine similarity in Sentence-BERT space) confirms that the collected annotations are consistent and of high quality. ExDDV comes with an official split into training, validation and test, which facilities reproducibility of results and future comparisons. 

We further evaluate a number of vision-language models (VLMs) on ExDDV, comparing various architectures, training procedures and supervision signals. In terms of architectures, we experiment with BLIP-2 \cite{li-ICML-2023}, Phi-3-Vision \cite{abdin-arxiv-2024} and LLaVA-1.5 \cite{liu-NeurIPS-2023}. In terms of training strategies, we consider pre-trained versions, as well as versions based on in-context learning and fine-tuning. Regarding the supervision signals, we consider text descriptions alone or in combination with clicks. For click supervision, we study two alternative approaches, namely soft and hard input masking. Our empirical results show that fine-tuning provides the most accurate explanations for all VLMs, confirming the utility of ExDDV in developing robust explainable models for deepfake videos. Moreover, we find that both text and click supervision signals are required to jointly localize and describe the observed artifacts, as well as to generate top-scoring explanations.

In summary, our contribution is threefold:
\begin{itemize}
    \item We introduce the first dataset for explainable deepfake detection in video, comprising 5.4K videos that are manually labeled with descriptions, clicks and difficulty levels.
    \item We study various VLM architectures and training strategies for explainable deepfake detection, all leading to a comprehensive benchmark.
    \item We publicly release our dataset and code to reproduce the results and foster future research.
\end{itemize}

\section{Related Work}

\noindent
\textbf{Deepfake detection in video.}
Nowadays, deepfakes have started to pose a real threat, since generative methods have significantly evolved and their number has increased. Nevertheless, substantial effort has been made to develop detection methods \cite{croitoru-arXiv-2024} and counter the misuse of generative AI technology. Early methods for deepfake detection in video were based on convolutional networks \cite{agarwal-WIFS-2020, gu-AAAI-2022, guera-AVSS-2018, sabir-CVPR-2019, amerini-IHMMSec-2020, masi-ECCV-2020, montserrat-CVPR-2020, hu-AAAI-2022, liu-WACV-2023, zhao-ICICS-2020, cozzolino-ICCV-2021, gu-ECCV-2022}. To handle both spatial and temporal dimensions, two different strategies are commonly adopted. The first is to apply 2D convolutions on individual frames and subsequently combine the resulting latent representations either by using basic operations (such as pooling or concatenation) \cite{agarwal-WIFS-2020, gu-AAAI-2022} or by employing recurrent neural networks \cite{guera-AVSS-2018, sabir-CVPR-2019, amerini-IHMMSec-2020, masi-ECCV-2020, montserrat-CVPR-2020, hu-AAAI-2022, liu-WACV-2023}. The second strategy involves extending the convolutions to 3D to capture spatio-temporal features \cite{zhao-ICICS-2020, cozzolino-ICCV-2021, gu-ECCV-2022}.

To learn more robust feature representations or enhance the detection of frame-level inconsistencies and motion artifacts, researchers have incorporated attention mechanisms into deepfake detectors \cite{bonettini-ICPR-2021, wang-AAAI-2023, yan-arXiv-2024}. Some recent works \cite{zheng-ICCV-2021, guan-neurIPS-2022, choi-CVPR-2024} employed the transformer architecture \cite{vaswani-NeurIPS-2017, dosovitskiy-ICLR-2021}. Such methods have a superior ability to capture long-range dependencies and are effectively applied to detect temporal inconsistencies.

To the best of our knowledge, current video deepfake detectors do not have intrinsic capabilities to explain their decisions. This is primarily caused by the lack of deepfake datasets providing explanatory annotations for the video content.

\noindent
\textbf{Deepfake video datasets.} The task of deepfake detection has been extensively studied, and thus, there are many datasets that are now publicly available. Among these, the most notable are LAV-DF \cite{cai-DICTA-2022}, GenVideo \cite{chen-arXiv-2024}, DeepFake Detection Challenge \cite{dolhansky-arXiv-2020}, DeeperForensics \cite{jiang-CVPR-2020}, FakeAVCeleb \cite{khalid-neurIPS-2021}, Celeb-DF \cite{li-CVPR-2020}, FaceForensics++ \cite{rossler-ICCV-2019}, WildDeepfake \cite{zi-ACMMM-2020} and MAVOS-DD \cite{Croitoru-Arxiv-2025}. Although such datasets contain the binary label (real or fake) associated with each video, they do not provide other kinds of annotations about the video content. Unlike existing datasets, we provide a dataset for deepfake detection with textual explanations for the artifacts observed by human annotators, along with clicks (points) that indicate artifact locations. To the best of our knowledge, our novel dataset is the first to provide explanatory annotations for deepfake video content.

\noindent
\textbf{Explainable deepfake detection.} The research community has extensively explored explainable AI (XAI), proposing various approaches \cite{lei-EMNLP-2016, zhong-arXiv-2019, jain-ACL-2020, mohankumar-ACL-2020, liu-ICML-2022,ribeiro-SIGKDD-2016,scott-NeurIPS-2017,selvaraju-IJCV-2020}. While there are many established methods for XAI, such as Gradient-weighted Class Activation Mapping (Grad-CAM) \cite{selvaraju-IJCV-2020}, SHapley Additive exPlanations (SHAP) \cite{scott-NeurIPS-2017} or LIME \cite{ribeiro-SIGKDD-2016}, explainable AI has barely been applied to deepfake detection. \citet{ishrak-arxiv-2024} implemented a binary classification model to detect whether video frames are artificially generated or not. Then, they employed Grad-CAM \cite{selvaraju-IJCV-2020} to estimate the salient regions that could explain the prediction, subsequently verifying if these regions overlap with the area of the face. Grad-CAM is a generic framework that back-propagates the gradients at various layers and computes a global average to obtain saliency maps.

Other methods opted for detecting deepfakes focusing only on a specific factor. Using only convolutional-based architectures, \citet{haliassos-CVPR-2021, haliassos-CVPR-2022} designed a detection method that concentrates on mouth movements, while \citet{demir-WACV-2024} proposed an approach that analyzes face motions. Due to the focus on specific factors, these methods can inherently provide some limited level of explainability.

To the best of our knowledge, none of the existing explainable AI methods are specifically designed to explain deepfake videos. However, we acknowledge that researchers have studied the explainability of generated images. For example, the WHOOPS dataset \cite{bitton-ICCV-2023} provides explanations for why synthetic images defy common sense. The authors note that such images can be easily identified by humans, raising the question if machines can do the same. In contrast, we focus on deepfake video content instead of generated images. We emphasize that deepfake content is potentially harmful to humans. Since deepfake videos are typically generated with a harmful intent in mind, they are not aimed to defy common sense (on the contrary). We thus consider the study of \citet{bitton-ICCV-2023} as a complementary work to our own.

\begin{figure}[!t]
    \centering
    \includegraphics[width=1\linewidth]{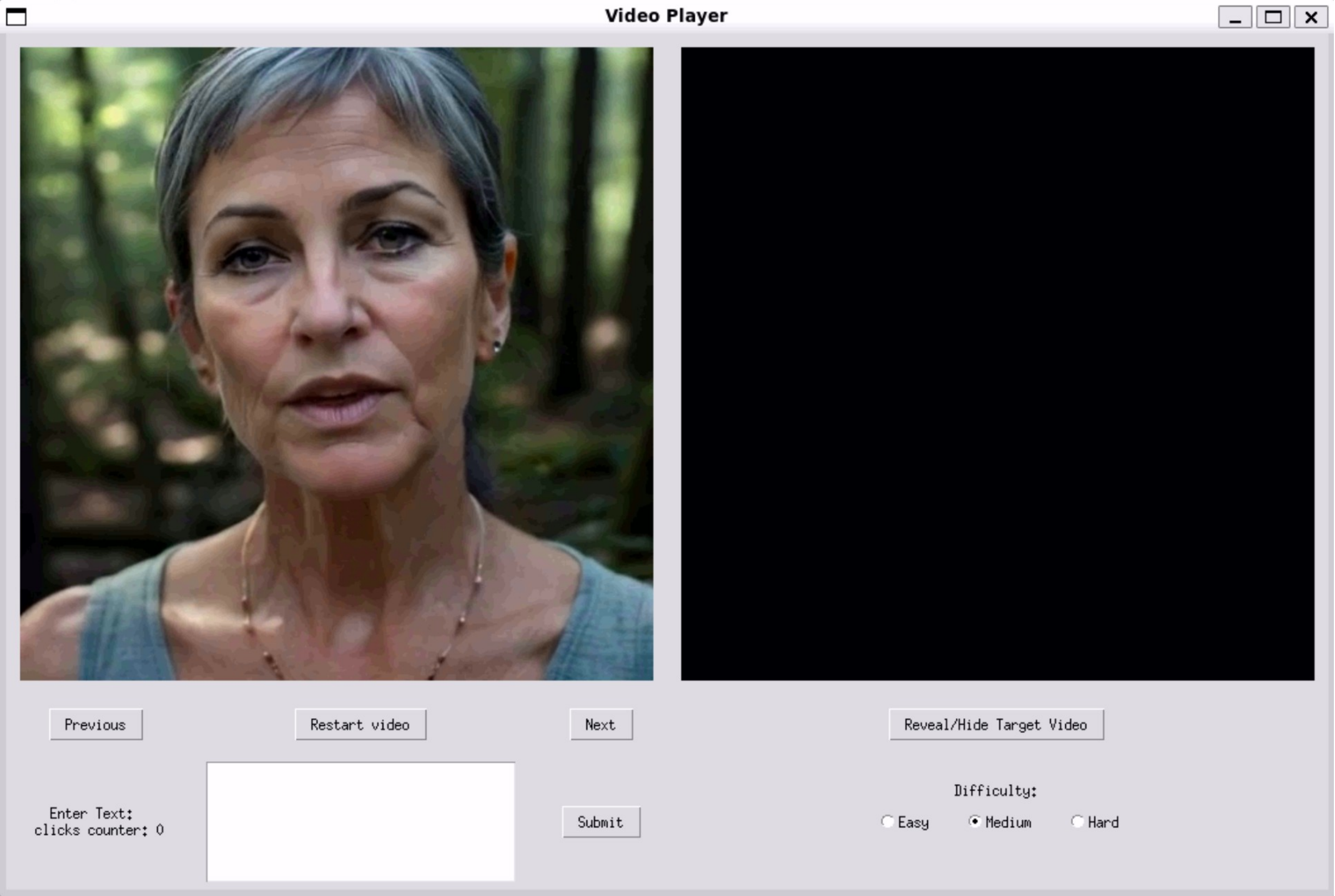}
    \vspace{-0.65cm}
    \caption{A screenshot of the application used to annotate ExDDV.}
    \label{fig:gui}
    \vspace{-0.4cm}
\end{figure}

\section{Proposed Benchmark}

The main contribution of our work is to introduce ExDDV, a benchmark specifically designed to facilitate human-interpretable explanations of deepfake videos. The novelty of ExDDV stems not only from being the first of its kind, but also from its comprehensive exploration of the task.

\noindent
\textbf{Video collection.} The fake videos are collected from four different sources: DeeperForensics \cite{jiang-CVPR-2020}, FaceForensics++ \cite{rossler-ICCV-2019}, DeepFake Detection Challenge (DFDC) \cite{dolhansky-arXiv-2020} and BioDeepAV \cite{croitoru-arXiv-2024}. This ensures that the dataset covers a wide range of generative methods and includes various video resolutions and durations. As real video samples, we use the original movies from DeeperForensics and FaceForensics++.
From FaceForensics++, we include all videos generated by the Face2Face, FaceSwap, and FaceShifter methods. From the other two datasets (DFDC and BioDeepAV), we randomly select a number of videos. In \cref{tab:dataset}, we present the breakdown of source datasets from which we gather videos for ExDDV. We provide an official split of the video collection into 4,380 training videos, 482 validation videos and 485 test videos.

\begin{table}[t]
\setlength\tabcolsep{0.4em}
  \centering
  \begin{tabular}{|l|l|c|}
    \hline
    \textbf{Source Dataset} & \textbf{Method} & \textbf{\#Samples} \\
    \hline
    \hline
    DeeperForensics & \multirow{2}{*}{Real} & \multirow{2}{*}{1,000} \\
    FaceForensics++ & & \\
    \hline
    DeeperForensics & DF-VAE & 1,000 \\
    \hline
    \multirow{3}{*}{FaceForensics++} 
                & Face2Face   & 1,000 \\
    \cline{2-3}
                & FaceSwap    & 1,000 \\
    \cline{2-3}
                & FaceShifter & 1,000 \\
    \hline
    DeepFake Detection & \multirow{2}{*}{Multiple} & \multirow{2}{*}{269} \\
    Challenge &  &  \\

    \hline
    BioDeepAV &  Talking-face & 100 \\
    \hline
    \textbf{Total} & & \textbf{5,369}\\
    \hline
  \end{tabular}
    \vspace{-0.2cm}
  \caption{Source datasets and generative methods for the deepfake videos included in ExDDV. Our dataset contains 1,000 real videos and 4,369 fake videos generated by various methods.}
  \label{tab:dataset}
  \vspace{-0.1cm}
\end{table}

\begin{table}[t]
\centering
\begin{tabular}{|l|c|}
\hline
\textbf{Measure} & \textbf{Average Score} \\
\hline\hline
Sentence-BERT & 0.6238 \\
BERTScore & 0.3857 \\
BLEU & 0.0808 \\
METEOR  & 0.2349 \\
ROUGE-1 & 0.3023 \\
ROUGE-2 & 0.1037 \\
ROUGE-L & 0.2434 \\
\hline
\end{tabular}
\vspace{-0.2cm}
\caption{Inter-annotator agreement among textual descriptions. Sentence-BERT and BERTScore indicate semantic alignment, while BLEU, METEOR, and ROUGE capture n-gram overlaps. Higher scores indicate a better alignment.}
\label{tab:text-metrics}
\vspace{-0.4cm}
\end{table}

\noindent
\textbf{Annotation.}
To annotate the dataset, we developed a simple application from scratch, with a custom graphical user interface (GUI), which is shown in \cref{fig:gui}. 
The annotation efforts are carried out by two paid human annotators, who are experts in the field. Each dataset is equally split in two subsets, such that each method and each possible kind of visual issue is described from two distinct points of view. More information about the annotation process can be found in the supplementary.
A couple of annotated examples are shown in \cref{fig:samples1} and \cref{fig:examples_extended}. 

\noindent
\textbf{Post-processing.}
We post-process the text annotations to correct spelling errors. We employ an automatic spell checker \cite{pyspellchecker} to find the misspelled words, and then manually correct them. We also change the forms of some words in order to consistently use American English instead of British English (\eg changing ``colour'' to ``color'').


\noindent
\textbf{Inter-annotator agreement.}
To estimate the inter-annotator agreement and guarantee the consistency of the annotations, we provide 100 deepfake videos to both annotators. All these videos have explanations and clicks from both humans.

We first evaluate the similarity between the text descriptions provided by the two annotators in terms of two semantic similarity metrics based on Sentence-BERT \cite{reimers-EMNLP-2019} and BERTScore \cite{zhang-ICLR-2020}, respectively. We hereby emphasize that conventional scores, such as BLEU \cite{Papineni-ACL-2002}, are not able to capture if two sentences have the same meaning when they contain different words \cite{Callison-Burch-EACL-2006}, \eg annotators use synonyms to refer to the same concept.
We therefore consider the similarities based on Sentence-BERT \cite{reimers-EMNLP-2019} and BERTScore \cite{zhang-ICLR-2020} as more reliable. The results shown in \cref{tab:text-metrics} demonstrate a close semantic alignment between textual descriptions, with an average cosine similarity in the Sentence-BERT embedding space of $0.6238$. The discrepancy between the higher Sentence-BERT and BERTScore values and the lower BLEU, METEOR and ROUGE scores indicates that the annotators rather used different words to explain the same artifacts in many cases.

\begin{table}[t]
\centering
\setlength\tabcolsep{0.5em}
\begin{tabular}{|c|c|c|c|c|}
\hline
\textbf{Temporal} & \textbf{Spatial} & \multicolumn{3}{c|}{\textbf{Agreement (\%)}} \\
\cline{3-5}
\textbf{Window} & \textbf{Window} & \textbf{Temporal} & \textbf{Spatial} & \textbf{Joint} \\
\hline
\hline
30 & 50$\times$50  & 54.70& 79.60& 35.10 \\
60 & 75$\times$75  & 66.14& 89.69& 53.82 \\
120 & 100$\times$100 & 84.54& 93.94& 75.87 \\
\hline
\end{tabular}
\vspace{-0.2cm}
\caption{Spatial and temporal inter-annotator agreement for the provided clicks. We report the percentage of clicks that match in specific temporal, spatial and joint (spatial-temporal) windows. Higher percentage points indicate a better agreement.}
\label{tab:click_agreement_joint}
\vspace{-0.1cm}
\end{table}

We also assess the temporal and spatial agreement for the clicks provided by the two annotators. We measure the temporal alignment in terms of the percentage of click pairs that match inside a temporal window of at most 30, 60 and 120 frames, respectively. Since the average FPS is 30, the window lengths correspond to 1, 2 and 4 seconds, respectively. Similarly, we measure the spatial alignment in terms of the percentage of click pairs that match inside a spatial window of 50, 75, and 100 pixels. Finally, we compute the joint spatio-temporal agreement as the percentage of click pairs situated inside the same temporal and spatial windows. As shown in \cref{tab:click_agreement_joint}, more than $53\%$ of the clicks made by the annotators are paired within a reasonable spatio-temporal window of 60 frames (about $2$ seconds) and $75\times75$ pixels. This high percentage of matching clicks indicates a significant agreement in terms of the artifact locations pointed out by the two annotators through clicks.


Overall, the various inter-annotator agreement measures indicate that the annotations are consistent. In addition, we also visually inspected the collected annotations and confirmed that they are of high quality. 

\begin{table}[t]
\centering
\begin{tabular}{|l|c|c|c|}
\hline
\textbf{Statistic} & \textbf{Min} & \textbf{Max} & \textbf{Average} \\
\hline\hline
FPS  & 8 & 60 & 27.66 \\
Frames & 50 & 1814 & 477.5 \\
Length (seconds) & 2.01 & 72.56 & 17.54 \\
Clicks & 1 & 35 & 4.9 \\
\hline
\end{tabular}
\vspace{-0.2cm}
\caption{Summary of statistics for ExDDV.}
\label{tab:stats}
\vspace{-0.4cm}
\end{table}

\noindent
\textbf{Statistics.}
The dataset consists of a total of 5,369 videos and 2,553,148 frames, with an average number of frames per video of approximately 477. We report several statistics about ExDDV in \cref{tab:stats}. The frames per second (FPS) rate varies from $8$ to $60$, $90\%$ of all videos having either $25$ or $30$ FPS. The total number of clicks is 21,282. On average, each movie is annotated with around $4.9$ clicks, the number of clicks per video ranging from $1$ to $35$. ExDDV comprises a wide range of video resolutions, from $480\times272$ to $1080\times1920$ pixels, with most videos having $720\times1280$, $480\times640$ or $1080\times1920$ pixels (see \cref{fig:bar_resolutions} from the supplementary for the number of videos per resolution). Each video is annotated with a single text description. 

\section{Explainable Methods}

We consider various vision-language models as candidates for deepfake video XAI methods. Although there have been many efforts on visual question answering (VQA) \cite{bigham-UIST-2010, malinowski-NeurIPS-2014, antol-ICCV-2015, gao-NeurIPS-2015, malinowski-CVPR-2015, ren-NeurIPS-2015, ma-AAAI-2016, yang-CVPR-2016, piergiovanni-arXiv-2022}, it was only with the significant growth of large language models (LLMs) \cite{radford-openai-2018, ouyang-NeurIPS-2022, naveed-arXiv-2023, touvron-arXiv-2023, jiang-arXiv-2023, minaee-arXiv-2024, liu-arXiv-2024} that capable vision-language models (VLMs) have recently emerged. For our task, we consider three different model families: Bootstrapping Language-Image Pre-training (BLIP) \cite{li-ICML-2022, li-ICML-2023}, Large Language and Vision Assistant (LLaVA) \cite{liu-NeurIPS-2023} and Phi-3 \cite{abdin-arxiv-2024}.

\subsection{Baseline architectures}

All chosen VLMs involve three components: an image encoder, a text encoder, and a decoder. First, the input image and the query are encoded with their corresponding modules. The second step is to project the visual tokens into the same vector space as the text tokens. Finally, all tokens are combined and decoded to generate an answer. 

\noindent
\textbf{BLIP-2.} The original BLIP architecture is designed to train a unified VLM end-to-end. It bootstraps from noisy image-text pairs by generating synthetic captions and filtering out noise. The training regime combines contrastive learning, image-text matching and autoregressive language modeling. 
BLIP-2 avoids end-to-end learning by taking a different route. More specifically, it introduces a lightweight Querying Transformer (Q-Former) that serves as a bridge between the image encoder and the LLM. The Q-Former employs a fixed set of learnable query tokens to extract the most relevant features from the image encoder, resulting in an efficient design with minimal computational costs. 
We opt for the BLIP-2 model based on an Open Pretrained Transformer (OPT) backbone with 2.7B parameters. For efficiency, we use the version optimized through \texttt{int8} quantization.

\noindent
\textbf{Phi-3-Vision.}
The Phi-3 \cite{abdin-arxiv-2024} suite of models ranges in functionality from solving pure language tasks to multimodal (vision and language) tasks. For our task, we employ the {Phi-3-Vision} model, which adopts a data-centric approach rather than simply increasing model size or applying distinct training regimes. The model is pre-trained using a curated pipeline that combines heavily filtered public web data with synthetic data generated by LLMs. This pre-training strategy first develops a broad language understanding, then enhances the model's reasoning and logical inference capabilities. We select the architecture based on 128K tokens, which uses flash attention.

\noindent
\textbf{LLaVA-1.5.} The LLaVA \cite{liu-NeurIPS-2023} family seamlessly integrates visual inputs with an LLM to generate natural language descriptions for visual tasks. When an image is provided as input, a pre-trained vision encoder extracts the relevant features. These are further projected into the LLM's token space through a linear transformation that preserves both spatial and semantic information. The LLM processes this sequence auto-regressively, producing responses that are inherently grounded in visual input. In our application, this approach enables the model not only to accurately answer questions about deepfake detection, but also to provide detailed and interpretable explanations of its reasoning. We opt for the LLaVA-1.5 version, with 7B parameters.

\begin{figure}[!t]
    \centering
    \includegraphics[width=1\linewidth]{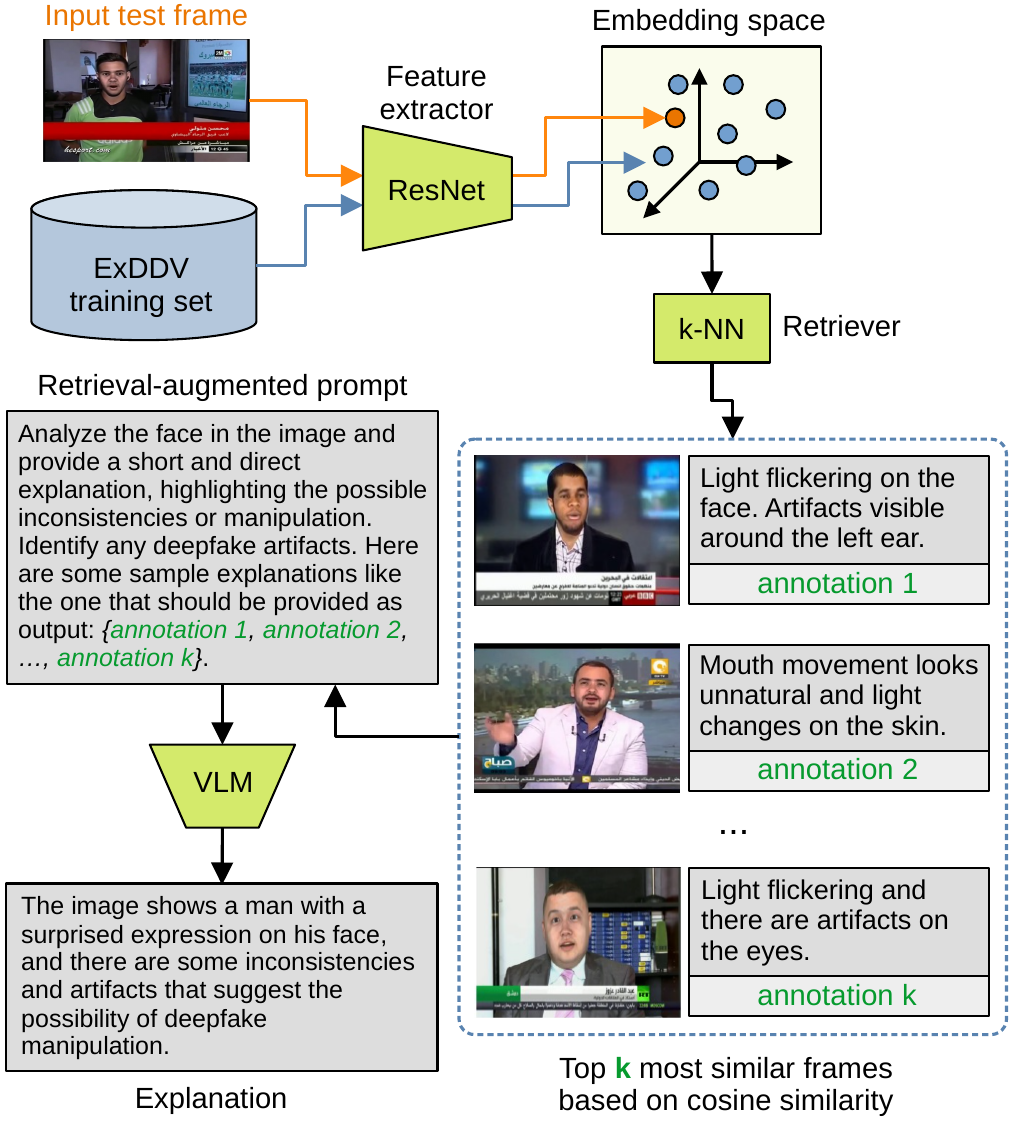}
    \vspace{-0.65cm}
    \caption{Overview of the in-context learning pipeline, which retrieves deepfake annotations from visually similar training frames using a k-NN based on a ResNet backbone. Best viewed in color.}
    \label{fig:incontext}
    \vspace{-0.2cm}
\end{figure}

\noindent
\textbf{Prompting.}
We prompt the pre-trained VLMs with the following query to obtain explanations about visual artifacts: \emph{``Analyze the face in the image. Identify any deepfake artifacts (if any), focusing specifically on the affected parts of the face. Provide a short and direct explanation highlighting the inconsistencies or manipulations''}.

\subsection{In-context learning}

\noindent
\textbf{Pipeline.}
In \cref{fig:incontext}, we illustrate our few-shot in-context learning pipeline. 
The training examples are chosen by a k-nearest neighbor (k-NN) model for each test video. The k-NN model extracts feature vectors with the CLIP \cite{radford-PmLR-2021} image encoder, which is based on the ResNet \cite{he-CVPR-2016} architecture. Features are only extracted from the training frames that are annotated with clicks. The frames are further used independently by the k-NN. For efficient inference, we store the training feature embeddings along with the associated annotations. When a test image is provided, we obtain its features via a similar process, extracting the ResNet features for the middle frame. Then, we search for the closest $k$ training frames based on the cosine similarity between feature embeddings. The annotations of the nearest neighbors are used to enrich the custom prompt given to the VLM. The k-NN procedure is supposed to provide a set of $k$ deepfake explanations that are potentially relevant for the test video. The custom prompt instructs the VLM to provide a similar explanation for the input test video.

\noindent
\textbf{Hyperparameters.} We consider two alternative CLIP image encoders for the k-NN: ResNet-50 and ResNet-101. For the number of neighbors $k$, we test values in the set $\{1, 3, 5, 10\}$. The best choice is $k=5$.  To perform the k-NN search at various representation levels, we extract features at three different depth levels. We use low-level features right before the first residual block, mid-level features from an intermediate residual block (equally distanced from the input and output), and high-level features from the last layer. The features producing the best results are the high-level ones. We keep the default parameters of all VLMs during inference, except for the temperature, which is set to $0$ to reduce the chance of generating hallucinations. 



\subsection{Fine-tuning}

\noindent
\textbf{Pipeline.}
On the one hand, we follow the default fine-tuning methodology for the BLIP-2 model. On the other hand, given the increased memory requirements of Phi-3-Vision and LLaVA-1.5, we employ Low-Rank Adaptation (LoRA) \cite{hu-ICLR-2022} to fine-tune these two models. 

To make a prediction for a whole video, the first step is to extract multiple frames from the video. Our strategy is to use equally-spaced frames between the first and last frames, while the number of frames depends on the model. Since BLIP-2 and Phi-3-Vision are lighter models, we provide 5 frames as input to these models. Phi-3-V supports multiple frames, while for BLIP-2, we process each frame with the visual encoder, then aggregate the resulting embeddings and provide the average embedding to the Q-former.
Along with the input frames, we provide a unique query prompt: \emph{``What is wrong in the images? Explain why they look real or fake''}. In our preliminary experiments, we tried to vary the prompt, considering even more complex prompts. However, we did not observe any significant differences, so we decided to stick with a single and relatively simple prompt for the presented experiments.

\begin{figure}[!t]
    \centering
    \includegraphics[width=0.88\linewidth]{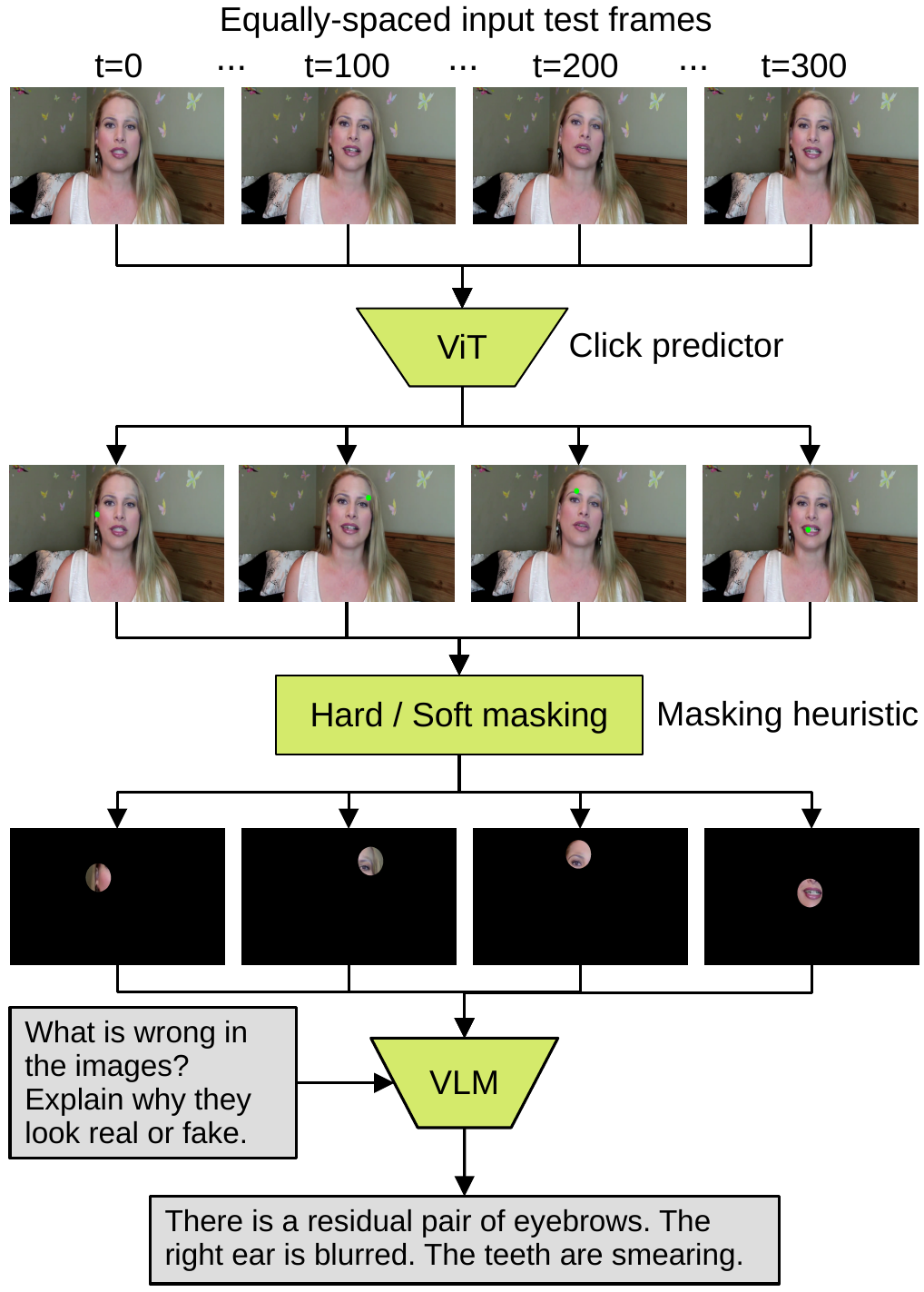}
    \vspace{-0.2cm}
    \caption{Our click supervision pipeline at inference time. A ViT-based click predictor estimates click coordinates for the input frames. A hard or soft masking is applied to mask the area outside the region of interest. The masked frames are given as input to a fine-tuned VLM. Best viewed in color.}
    \label{fig:click_supervision}
    \vspace{-0.2cm}
\end{figure}

\noindent
\textbf{Hyperparameters.}
Phi-3-Vision and LLaVA are trained for 10 epochs with mini-batches of $16$ and $32$ samples, respectively. BLIP-2 is also fine-tuned for 10 epochs with a mini-batch size of $16$, but with a gradient accumulation of $2$. All models are optimized with AdamW, using a learning rate of $2\cdot 10^{-4}$ and a cosine annealing scheduler to reduce the learning rate during training. For Phi-3-V, we use a LoRA rank of $64$ and a dropout rate of $0.05$. For LLaVA, we set the rank to $128$ and do not employ dropout. The parameter $\alpha$ of LoRA is set to $256$ for both models.

\subsection{Click supervision}

We harness the additional supervision signal in our dataset, namely the locations of the visual flaws represented by mouse clicks. While this idea was found to be useful in a number of vision tasks \cite{papadopoulos-CVPR-2017, benenson-CVPR-2019, chen-ECCV-2022, cheng-CVPR-2022, luo-CVPR-2024}, to our knowledge, it has not been explored to train explainable video models. The rationale behind using click supervision is to guide the explainable model towards focusing on the region of interest (ROI) and consequently provide precise explanations.

In \cref{fig:click_supervision}, we present our pipeline based on click-supervision, which comprises a click predictor based on ResNet, and a frame masking heuristic, which preserves the ROI, while masking the rest of the frame. To train the click predictor, we first extract all the training frames in which a visual glitch is annotated through a click, along with the coordinates of each click. The coordinates are further normalized between 0 and 1 to make the click predictor invariant to distinct frame resolutions. Next, we train a ViT-based regression model to predict click coordinates. The ViT \cite{dosovitskiy-ICLR-2021} backbone is pre-trained on ImageNet. During inference, we apply the regressor on each test frame to predict the clicks. Then, we mask the image area outside the ROI, which is defined as a round area of radius $r$ around the click coordinates. We consider two alternative masking operations, soft and hard. Hard masking implies replacing the masked pixels with zero. Soft masking is based on a 2D Gaussian distribution centered in the predicted click location, where the $\sigma=r$. The masking operation is performed after the input images are normalized. 

\noindent
\textbf{Hyperparameters.} We consider two models pre-trained on ImageNet to predict clicks, a ViT-B and a ResNet-50. We replace their classification heads with a regression head to predict the two coordinates of a click. The click predictors are trained for $15$ epochs with the AdamW optimizer and a learning rate of $9 \cdot 10^{-5}$. The learning rate is reduced on plateau by half, if the validation loss does not change for two consecutive epochs. An important hyperparameter for the masking heuristic is the mask radius $r$, which needs to be fixed to provide just enough context for the VLM. We tuned $r$ between $50$ and $150$ pixels with a step of $5$ pixels. The optimal radius is $r=75$.

\section{Experiments and Results}
\label{sec:results}

\begin{table*}[t]
\centering
\setlength\tabcolsep{0.18em}
\begin{tabular}{|l|c|c|c|c|c|c|c|c|}
\hline
\textbf{Model} & \textbf{Masking} & \textbf{Sentence-BERT} & \textbf{BERTScore} & \textbf{BLEU} & \textbf{METEOR} & \textbf{ROUGE-1} & \textbf{ROUGE-2} & \textbf{ROUGE-L} \\
\hline
\hline
 BLIP-2 pre-trained & & $0.09$ & $0.03$ & $ 0.01$ & $0.04$ & $0.05$ & $0.00$ & $0.05$ \\
 \hline
 BLIP-2 in-context & & $0.44_{\pm0.008}$ & $0.19_{\pm0.004}$ & $0.03_{\pm0.001}$ & $0.21_{\pm0.001}$ & $0.14_{\pm0.003}$ & $0.06_{\pm0.001}$ & $0.14_{\pm0.001}$ \\

   BLIP-2 in-context & hard & $0.47_{\pm0.002}$ & $0.20_{\pm0.001}$ & $0.04_{\pm0.001}$ & $0.21_{\pm0.003}$ & $0.14_{\pm0.005}$ & $0.06_{\pm0.001}$ & $0.14_{\pm0.002}$ \\
   
  BLIP-2 in-context & soft & $0.46_{\pm0.004}$ & $0.20_{\pm0.003}$ & $0.04_{\pm0.001}$ & $0.20_{\pm0.001}$ & $0.14_{\pm0.001}$ & $0.06_{\pm0.001}$ & $0.14_{\pm0.002}$ \\
 \hline
BLIP-2 fine-tuned & & $0.45_{\pm0.009}$ & $0.29_{\pm0.002}$ & $0.09_{\pm0.010}$ & $0.14_{\pm0.015}$ & $0.21_{\pm0.010}$ & $0.09_{\pm0.007}$ & $0.20_{\pm0.009}$ \\

 BLIP-2 fine-tuned & hard & ${\color{blue}0.55}_{\pm0.005}$ & $\textcolor{blue}{0.36}_{\pm0.006}$ & $\textcolor{blue}{0.14}_{\pm0.005}$ & $\textcolor{blue}{0.22}_{\pm0.007}$ & $\textcolor{blue}{0.31}_{\pm0.004}$ & $\textcolor{blue}{0.38}_{\pm0.004}$ & $\textcolor{blue}{0.29}_{\pm0.004}$ \\
 
 BLIP-2 fine-tuned & soft & $0.54_{\pm0.006}$ & $\textcolor{blue}{0.36}_{\pm0.002}$ & $0.13_{\pm0.002}$ & $0.21_{\pm0.001}$ & $0.30_{\pm0.004}$ & $0.14_{\pm0.001}$ & $0.28_{\pm0.003}$ \\
\hline
\hline
 Phi-3-V pre-trained & & $0.25$ & $0.10$ & $0.01$ & $0.01$ & $0.05$ & $0.00$ & $0.04$ \\
\hline
 Phi-3-V in-context & & $0.30_{\pm0.002}$ & $0.14_{\pm0.003}$ & $0.03_{\pm0.002}$ & $0.18_{\pm0.005}$ & $0.15_{\pm0.002}$ & $0.05_{\pm0.002}$ & $0.13_{\pm0.002}$ \\
 
 Phi-3-V in-context & hard & $0.30_{\pm0.002}$ & $0.14_{\pm0.005}$ & $0.03_{\pm0.004}$ & $0.18_{\pm0.005}$ & $0.16_{\pm0.003}$ & $0.05_{\pm0.003}$ & $0.13_{\pm0.005}$ \\
 
Phi-3-V in-context & soft & $0.30_{\pm0.006}$ & $0.14_{\pm0.003}$ & $0.03_{\pm0.004}$ & $0.18_{\pm0.004}$ & $0.15_{\pm0.002}$ & $0.05_{\pm0.003}$ & $0.13_{\pm0.004}$ \\ 
 \hline
  Phi-3-V fine-tuned & & $0.42_{\pm0.005}$ & $0.30_{\pm0.002}$ & $0.06_{\pm0.001}$ & $0.20_{\pm0.004}$ & $0.21_{\pm0.005}$ & $0.07_{\pm0.005}$ & $0.19_{\pm0.006}$ \\
  
 Phi-3-V fine-tuned & hard & $\textcolor{blue}{0.53}_{\pm0.004}$ & $\textcolor{blue}{0.38}_{\pm0.003}$ & $\textcolor{blue}{0.09}_{\pm0.004}$ & $\textcolor{blue}{0.27}_{\pm0.004}$ & $\textcolor{blue}{0.28}_{\pm0.002}$ & $\textcolor{blue}{0.11}_{\pm0.004}$ & $\textcolor{blue}{0.25}_{\pm0.003}$ \\
 
 Phi-3-V fine-tuned & soft & $\textcolor{blue}{0.53}_{\pm0.003}$ & $\textcolor{blue}{0.38}_{\pm0.003}$ & $\textcolor{blue}{0.09}_{\pm0.008}$ & $0.26_{\pm0.011}$ & $\textcolor{blue}{0.28}_{\pm0.009}$ & $\textcolor{blue}{0.11}_{\pm0.010}$ & $\textcolor{blue}{0.25}_{\pm0.009}$ \\
\hline
\hline
 LLaVA pre-trained & & $0.39$ & $0.18$ & $0.02$ & $0.18$ & $0.12$ & $0.02$ & $0.10$ \\
 \hline
 LLaVA in-context & & $0.48_{\pm0.002}$ & $0.22_{\pm0.008}$ & $0.03_{\pm0.001}$ & $0.21_{\pm0.001}$ & $0.12_{\pm0.002}$ & $0.03_{\pm0.002}$ & $0.10_{\pm0.002}$ \\

 LLaVA in-context & hard & $0.47_{\pm0.002}$ & $0.29_{\pm0.010}$ & $0.05_{\pm0.001}$ & $0.22_{\pm0.001}$ & $0.20_{\pm0.002}$ & $0.04_{\pm0.000}$ & $0.16_{\pm0.002}$ \\
 
 LLaVA in-context & soft & $\textcolor{blue}{0.50}_{\pm0.006}$ & $0.23_{\pm0.004}$ & $0.04_{\pm0.001}$ & $0.20_{\pm0.002}$ & $0.12_{\pm0.002}$ & $0.04_{\pm0.001}$ & $0.11_{\pm0.002}$ \\
\hline
 LLaVA fine-tuned & & $0.45_{\pm0.003}$ & $0.33_{\pm0.003}$ & $0.07_{\pm0.002}$ & $0.23_{\pm0.003}$ & $0.24_{\pm0.001}$ & $0.08_{\pm0.001}$ & $0.21_{\pm0.001}$ \\
 LLaVA fine-tuned & hard & $0.49_{\pm0.011}$ & $\textcolor{blue}{0.35}_{\pm0.009}$ & $\textcolor{blue}{0.08}_{\pm0.005}$ & $0.24_{\pm0.009}$ & $\textcolor{blue}{0.25}_{\pm0.011}$ & $\textcolor{blue}{0.09}_{\pm0.007}$ & $\textcolor{blue}{0.22}_{\pm0.008}$ \\
 LLaVA fine-tuned & soft & $0.49_{\pm0.006}$ & $\textcolor{blue}{0.35}_{\pm0.004}$ & $\textcolor{blue}{0.08}_{\pm0.002}$ & $\textcolor{blue}{0.25}_{\pm0.002}$ & $\textcolor{blue}{0.25}_{\pm0.003}$ & $\textcolor{blue}{0.09}_{\pm0.000}$ & $\textcolor{blue}{0.22}_{\pm0.003}$ \\
\hline
\end{tabular}
\vspace{-0.2cm}
\caption{Results of various XAI models on ExDDV. Soft or hard input masking is based on click coordinates predicted by a ViT-based regressor. 
The top score for each model and evaluation metric is colored in {\color{blue}blue}.}
\label{tab:results}
\vspace{-0.2cm}
\end{table*}

\noindent
\textbf{Research questions.}
Through our experiments, we aim to address the following research questions (RQs):
\begin{enumerate}
    \item Are the collected annotations useful to train XAI models for deepfake videos?
    \item What is the performance impact of click supervision?
    \item How many data samples are required to train XAI models for deepfake videos?
    \item Can the locations of artifacts be accurately predicted?
\end{enumerate}

To address RQ1, we compare off-the-shelf (pre-trained) VLMs with VLMs based on two alternative training strategies, namely fine-tuning and in-context learning. To answer RQ2, in both types of learning frameworks, we harness the additional information provided through clicks and assess the performance gains brought by this supervision signal. To answer RQ3, we trained the top-scoring VLM with varying training set dimensions in the set $\{128, 256, 512, 1024, 2048, 4380 \}$. To address RQ4, we report results with two click predictors that are both fine-tuned on ExDDV training data.

\noindent
\textbf{Evaluation measures.} Although there are many measures to assess either the semantic similarity or the n-gram overlap of two text samples, such measures are not able to fully capture the similarity between texts. As a result, we evaluate the XAI models using a wide range of metrics, aiming to provide an extensive evaluation of the results. For semantic understanding, we employ Sentence-BERT \cite{reimers-EMNLP-2019} to embed both predicted and ground-truth descriptions, and compute the cosine similarity between the two. We also employ BERTScore \cite{zhang-ICLR-2020} to compute the similarity at the token level. Every token in the ground-truth is greedily matched with a token in the prediction to compute a recall. Similarly, every token in the prediction is matched with a token in the ground-truth to compute the precision. These are combined into an F1 score, called \emph{BERTScore}. To assess n-gram overlaps, we adopt the most popular evaluation measures used in image captioning: BLEU \cite{Papineni-ACL-2002}, METEOR \cite{banerjee-ACL-2005} and ROUGE \cite{lin-ACL-2004}. For these metrics, we set the maximum n-gram length to $n=2$. We evaluate click predictors in terms of the mean absolute error (MAE). We run each experiment three times and report the average scores and the standard deviations.


\noindent
\textbf{Quantitative results.}
The results of our experiments are shown in \cref{tab:results}. We present the results for all models (BLIP-2, Phi-3-Vision and LLaVA-1.5) and learning scenarios (pre-training, in-context learning and fine-tuning). Click supervision is integrated via soft or hard masking, respectively. Consistent with the inter-annotator agreement scores, we observe that all models yield better scores in terms of semantic measures than n-gram overlap measures. The gap can be explained by the fact that models generate varied outputs, often using alternative phrases and words to express the same concept.

\begin{figure}[!t]
    \centering
    \includegraphics[width=0.9\linewidth]{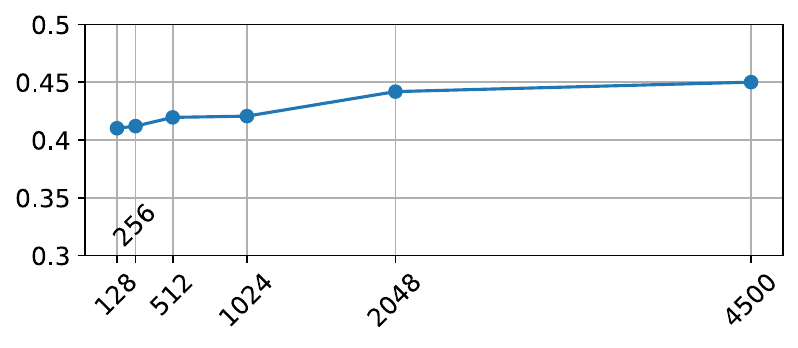}
    \vspace{-0.2cm}
    \caption{The performance of fine-tuned LLaVA (vertical axis) against the number of samples used for training (horizontal axis).}
    \label{fig:performance}
    \vspace{-0.3cm}
\end{figure}

\begin{figure*}[!t]
    \centering
    \includegraphics[width=1.0\linewidth]{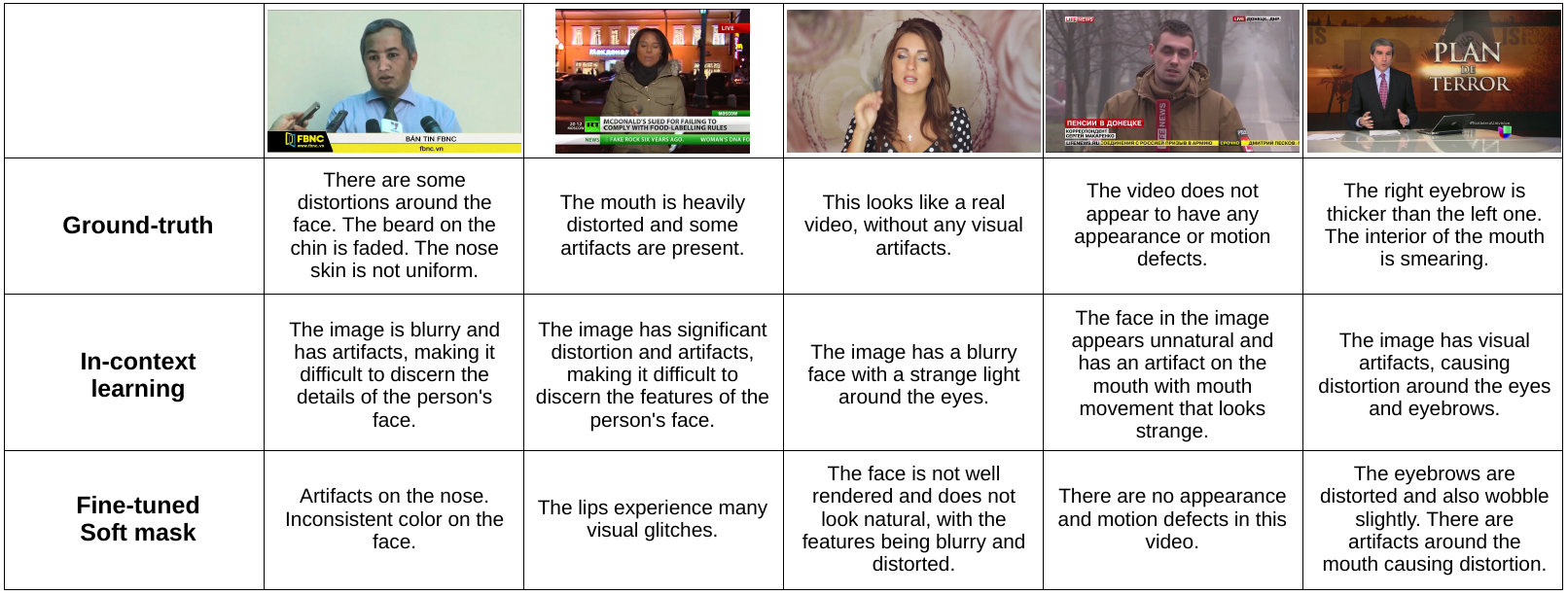}
    \vspace{-0.6cm}
    \caption{Qualitative examples for LLaVA based on in-context learning and LLaVA fine-tuned with soft input masking. Best viewed in color.} 
    \label{fig:xai_results_main}
    \vspace{-0.35cm}
\end{figure*}

All fine-tuned models surpass the pre-trained versions by large margins. Although the metrics indicate that the pre-trained LLaVA is close to the fine-tuned models, a visual inspection of its generated answers indicates that they are very generic, in many cases just describing the videos and not providing any information about their authenticity. The reported results offer strong evidence for RQ1, indicating that the collected annotations are useful to train XAI models, both via in-context learning and fine-tuning.

The reported scores also attest the advantages of integrating click predictions via soft and hard masking. This is observable for all three models, although the gains are somewhat lower for LLaVA. Both masking strategies appear to be equally effective. In response to RQ2, we find that using click supervision has a positive impact, boosting the performance of all tested VLMs. 


In \cref{fig:performance}, we showcase the performance of the fine-tuned LLaVA model for different dimensions of the training set. The performance reaches a plateau after 2,000 training samples. This observation suggests that our dataset contains enough samples to train an explainable deepfake detection model, elucidating RQ3.

In \cref{tab:keypoint_models}, we report the mean absolute errors for both ViT and ResNet-50 click predictors. The error of ViT is slightly lower, representing an average offset of only $12$ pixels w.r.t.~the ground-truth coordinates. These results confirm that the regression models can accurately localize visual artifacts, thus providing a positive answer to RQ4.

\begin{table}[!t]
\centering
\begin{tabular}{|l|c|c|}
\hline
\textbf{Model} & ViT & ResNet-50 \\
\hline\hline
\textbf{MAE}  & $0.0553$ & $0.0595$ \\
\hline
\end{tabular}
\vspace{-0.25cm}
\caption{Results of click predictors.}
\label{tab:keypoint_models}
\vspace{-0.1cm}
\end{table}

\noindent
\textbf{Qualitative results.}
Besides the quantitative measurements, we also present qualitative results. In \cref{fig:xai_results_main}, we illustrate some explanations provided by two variants of LLaVA. The examples include both relevant explanations as well as wrong explanations, \eg identifying artifacts on real videos. We present examples for additional versions of LLaVA in \cref{fig:xai_results} from the supplementary. 

\begin{figure}[t]
    \centering
    \includegraphics[width=1.0\linewidth]{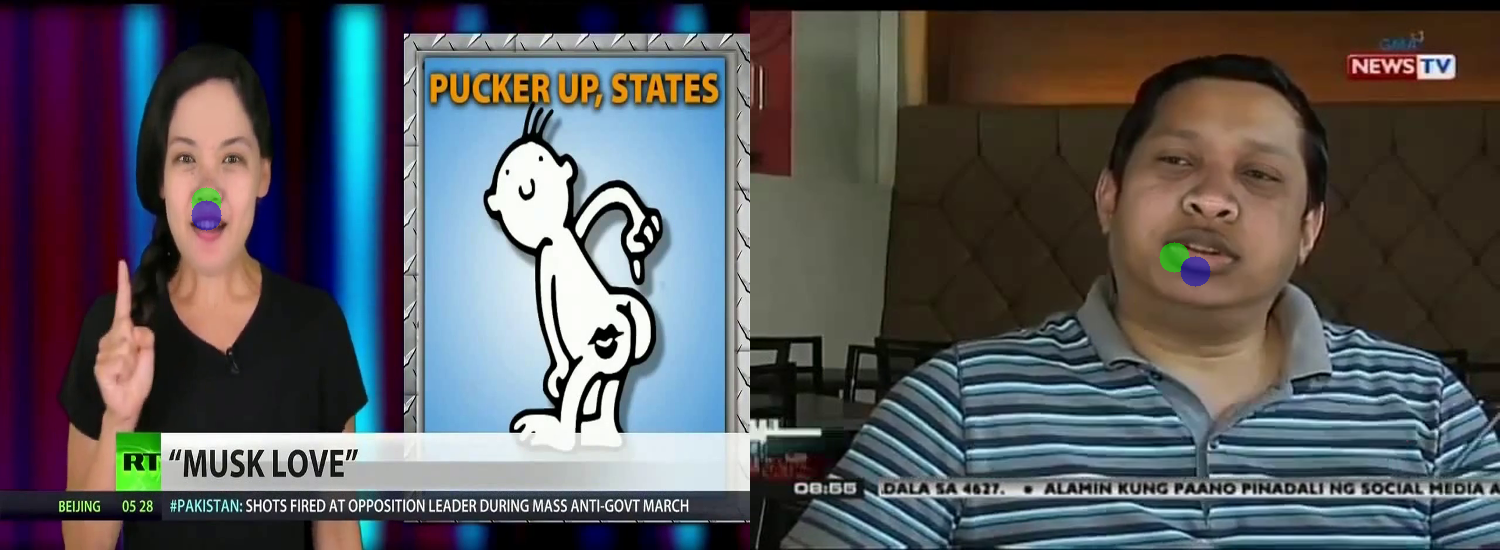}
    \vspace{-0.6cm}
    \caption{Qualitative examples of our click predictor. The green dots represent the ground-truth locations and the blue dots represent the corresponding predictions. Best viewed in color.}
    \label{fig:click_results_main}
    \vspace{-0.3cm}
\end{figure}

In \cref{fig:click_results_main}, we showcase some examples of how the ViT-based click predictor compares with the ground-truth click locations. We observe that the predictor is able to precisely locate visual artifacts. We present additional click prediction examples in \cref{fig:click_results} from the supplementary.

\section{Conclusion}
In our work, we introduced a novel dataset for explainable deepfake detection in videos and made it publicly available. ExDDV consists of 5.4K manually annotated real and fake videos. In addition to the explanations for each video, the dataset also contains the locations of visual artifacts. We also explored different VLMs on the explainable deepfake detection task and evaluated their performance. The empirical results showed that the models are capable of learning to predict the source of visual errors in fake videos, while also detecting real videos. While the reported results are promising, we found that the tested VLMs are well below the inter-annotator agreement scores, suggesting that further exploration is required to build more capable models. In future work, we also aim to harness the difficulty labels, \eg via curriculum learning, to boost performance.

With deepfake methods rapidly becoming more powerful and easier to access by the whole public, we consider our work as a stepping stone towards developing more robust and transparent detection models that will overcome the harms of deepfakes. We believe that our research will contribute to trustworthy AI systems that will only bring benefits to society, as well as reduce the skepticism around AI technology.

\section*{Acknowledgments}
This work was supported by a grant of the Ministry of Research, Innovation and Digitization, CCCDI - UEFISCDI, project number PN-IV-P6-6.3-SOL-2024-2-0227, within PNCDI IV. This research was also supported by the project ``Romanian Hub for Artificial Intelligence - HRIA'', Smart Growth, Digitization and Financial Instruments Program, 2021-2027, MySMIS no.~351416.

{
    \small
    \bibliographystyle{ieeenat_fullname}
    \bibliography{main}
}

\section{Annotation Process and Video Resolutions}
The GUI (shown in \cref{fig:gui}) contains two video players, side by side. The deepfake video is played on the left side of the window, while the corresponding real video is played on the right side. By default, the real video is not played. If the annotator needs to play the real video along with the deepfake one, they could simply press the button below the player screen. When the fake video is playing, the user can click anywhere on the frame to indicate the location of artifacts. Behind the scene, the application records the relative pixel location and the timestamp (\ie frame index) of the click. Under the fake video player, there is a text box in which the details describing the visual issues can be written by the user. In the bottom right, there are three radio buttons, which allow the user to indicate the difficulty level of identifying deepfake evidence. We ask users to label deepfake videos as \emph{hard}, when they need to play the deepfake video at least two times, or when they need to activate the real video to observe artifacts. In a similar manner, we instruct them to label videos as \emph{easy}, if they are able to identify more than one artifact with a single play of the deepfake video. For real videos, we do not collect clicks. 

\begin{figure}[th]
    \centering
    \includegraphics[width=0.975\linewidth]{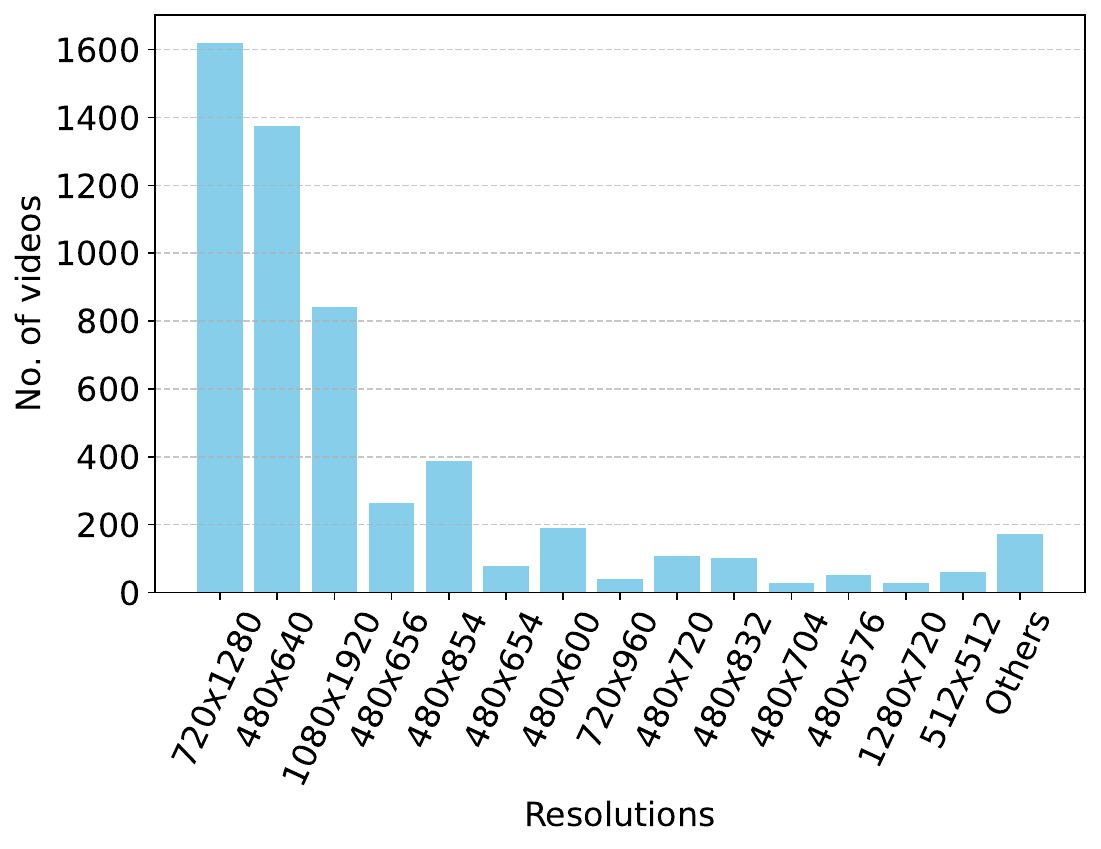}
        \vspace{-0.2cm}
    \caption{The number of videos for each frame resolution in ExDDV.}
    \label{fig:bar_resolutions}
    \vspace{-0.2cm}
\end{figure}

In Figure~\ref{fig:bar_resolutions}, we plot a bar chart showing the various resolutions that comprise ExDDV. The bar chart clearly shows that the first three resolutions are significantly more frequent than the others. 

\begin{figure*}[th]
    \centering
    \includegraphics[width=1.0\linewidth]{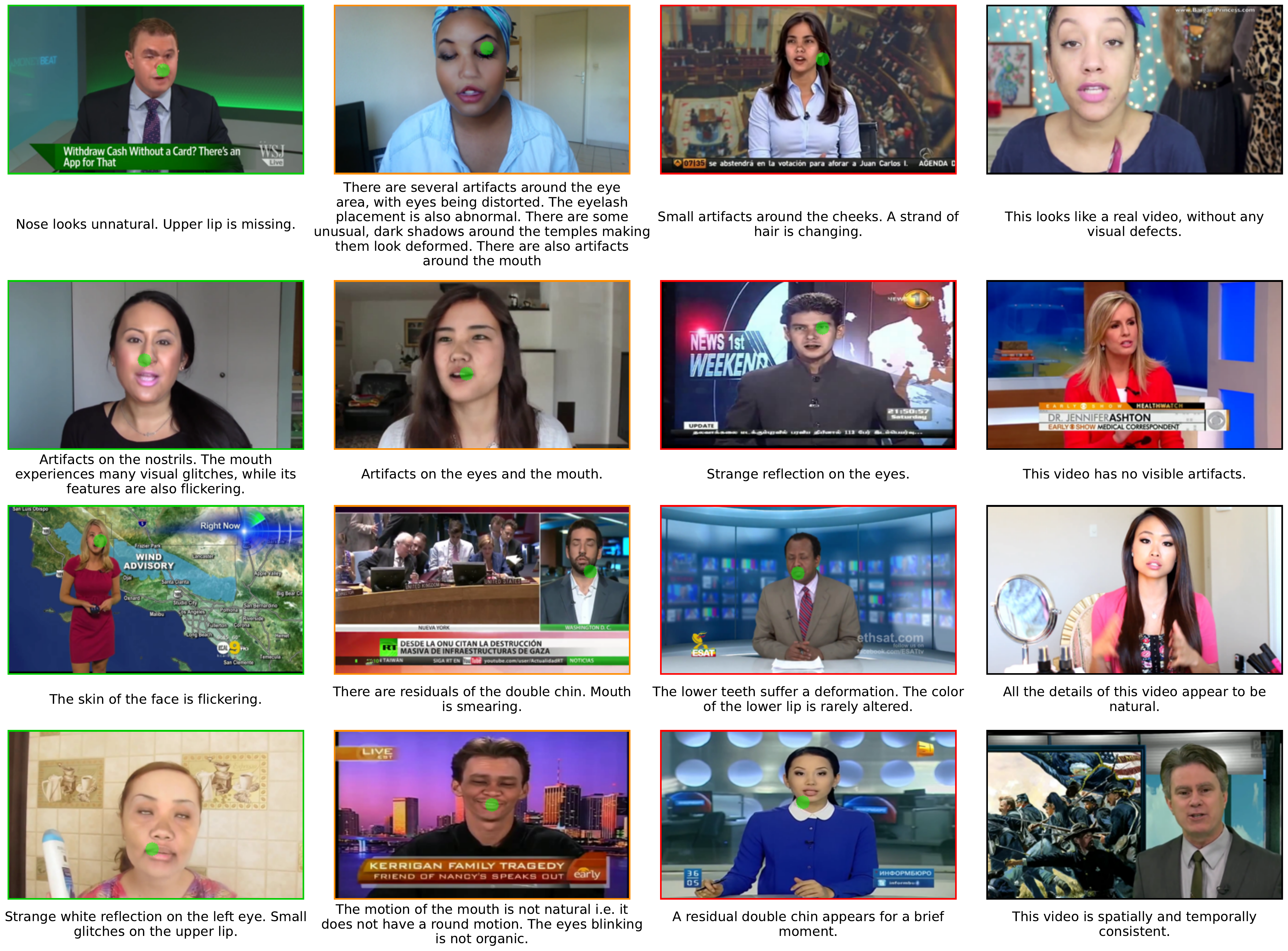}
    \caption{More examples of video frames from ExDDV with text and click annotations. Clicks are represented as large green dots. Real videos are not annotated with clicks or difficulty levels. The border color indicates the difficulty level: \textcolor{difficulty_green}{green}=easy, \textcolor{difficulty_orange}{orange}=medium, \textcolor{difficulty_red}{red}=hard, black=real.  Best viewed in color.}
    \label{fig:examples_extended}
\end{figure*}

In Figure~\ref{fig:examples_extended}, we present more annotated samples from ExDDV, having different levels of difficulty, click locations, and explanatory text lengths. 

\section{Additional Results}





\begin{figure*}
    \centering
    \includegraphics[width=1.0\linewidth]{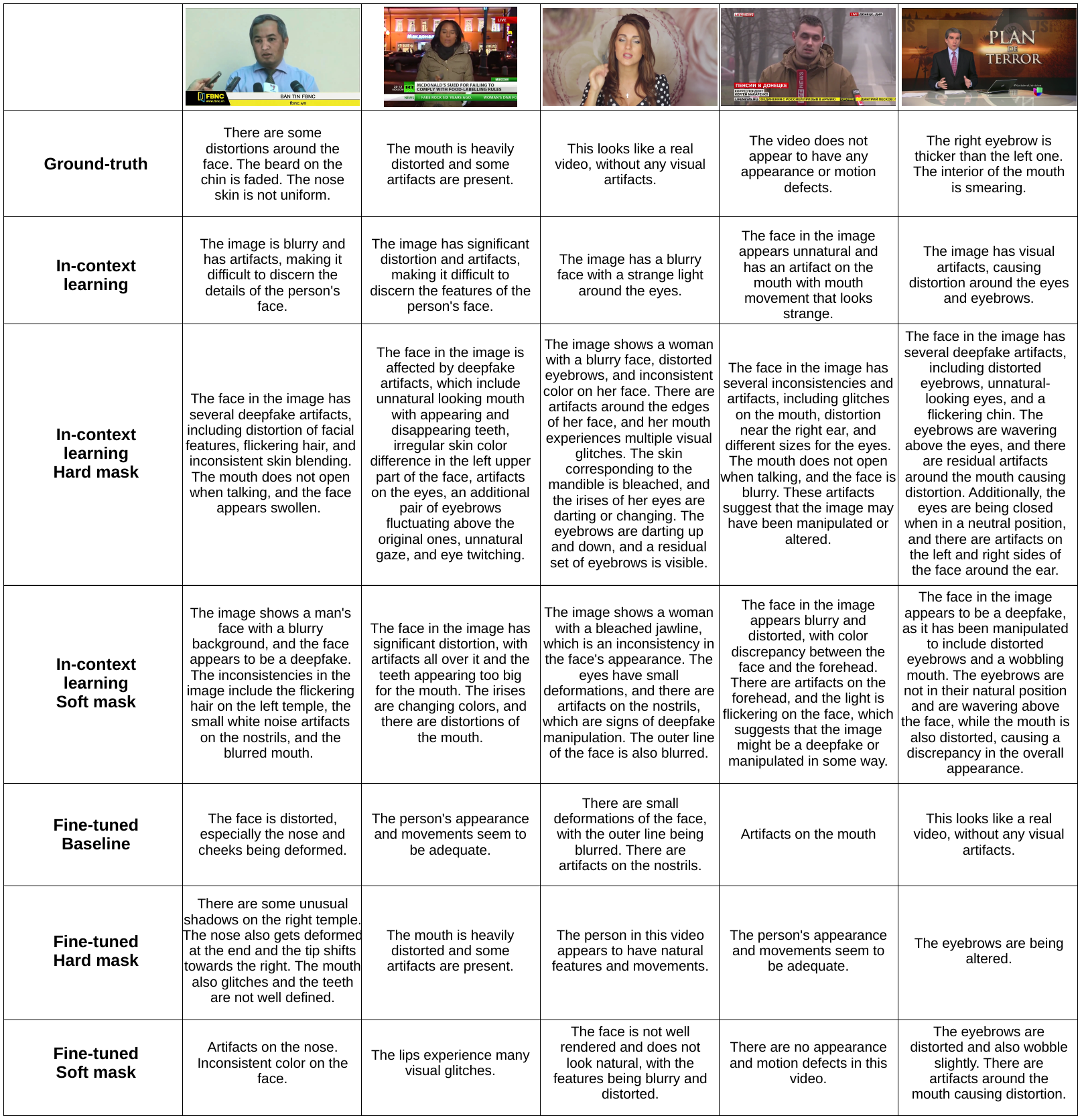}
    \caption{Qualitative examples for LLaVA in all training scenarios: pre-trained, in-context learning with and without masking, and fine-tuned with and without masking.}
    \label{fig:xai_results}
\end{figure*}

Figure~\ref{fig:xai_results} contains a comprehensive diagram with qualitative samples for all possible training scenarios applied on LLaVA \cite{liu-NeurIPS-2023}. The examples include both relevant explanations as well as wrong explanations, \eg identifying artifacts on real videos.

\begin{figure*}
    \centering
    \includegraphics[width=0.75\linewidth]{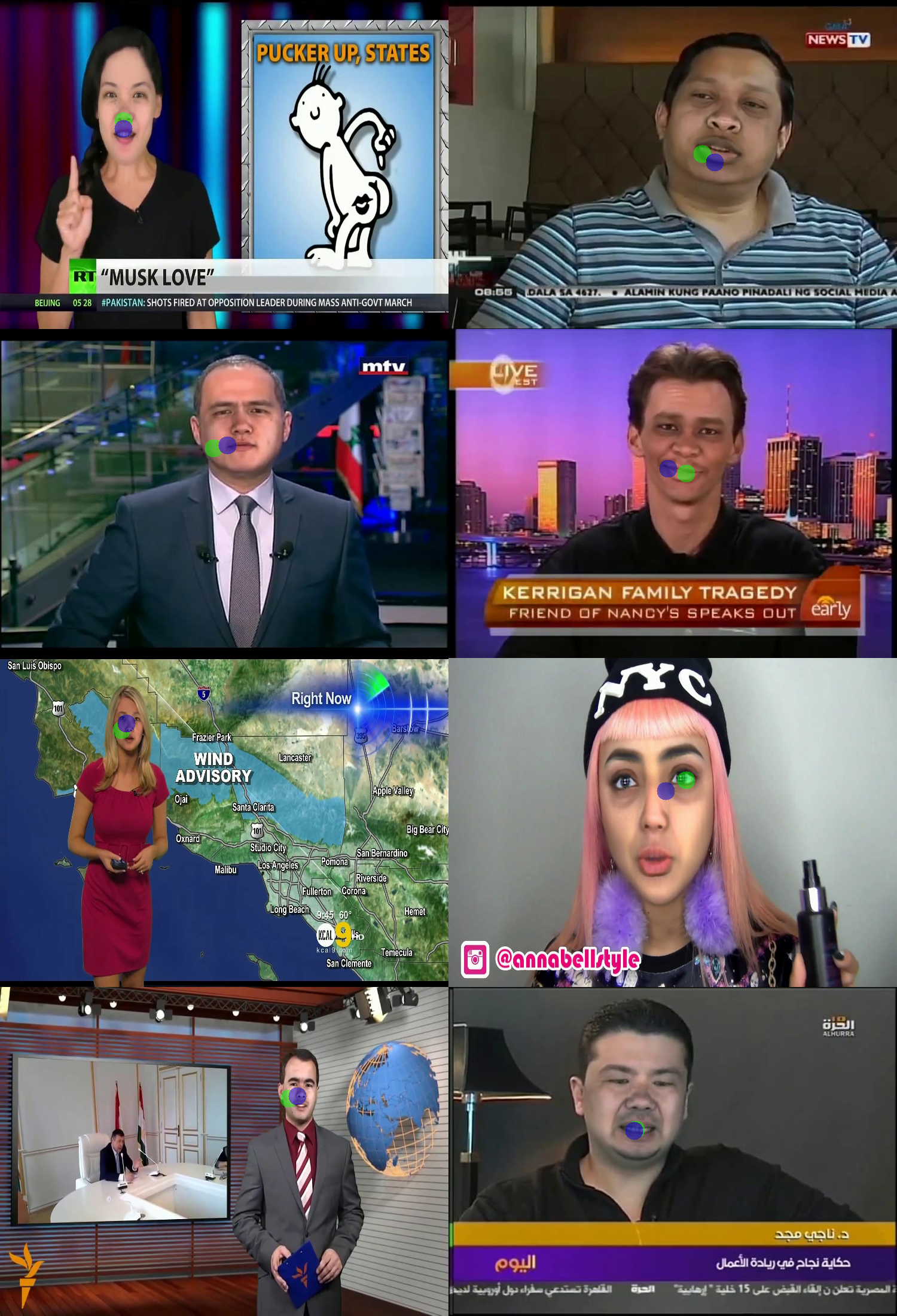}
    \caption{Qualitative examples of our click predictor. The green dots represent the ground-truth locations and the blue dots represent the corresponding predictions. Best viewed in color.}
    \label{fig:click_results}
\end{figure*}

In Figure~\ref{fig:click_results}, we showcase some examples of how the ViT-based click predictor compares with the ground-truth click locations. We observe that the predictor is able to precisely locate visual artifacts.

\section{Training Environments}
We have worked on multiple environments for our experiments. For the in-context learning experiments, we used a Tesla V100-SXM2 GPU with 32GB VRAM. 
Phi-3-Vision \cite{abdin-arxiv-2024} and LLaVA \cite{liu-NeurIPS-2023} were fine-tuned on a single H100 GPU with 80GB VRAM. BLIP-2 \cite{li-ICML-2023} was fine-tuned using an RTX 4090 GPU with 24GB VRAM. The same training environment as for BLIP-2 was used for the click predictors.

\end{document}